\documentclass[runningheads]{llncs}

\usepackage{eccv}

\usepackage{eccvabbrv}

\usepackage{graphicx}
\usepackage{booktabs}
\usepackage{soul}
\usepackage{multirow}
\usepackage{pifont}

\usepackage[accsupp]{axessibility}  % Improves PDF readability for those with disabilities.

\usepackage{hyperref}

\usepackage{orcidlink}

\usepackage{booktabs,tabularx,siunitx,xcolor,colortbl}
\definecolor{best}{RGB}{220,245,220}     % light green
\definecolor{second}{RGB}{235,240,255}   % light blue
\definecolor{best}{RGB}{220,245,220}     % light green
\definecolor{second}{RGB}{235,240,255}   % light blue
\newcommand{\best}[1]{\cellcolor{best}\textbf{#1}}
\newcommand{\secondbest}[1]{\cellcolor{second}\underline{#1}}
\newcommand{\src}[1]{\textsuperscript{\scriptsize #1}}
\newcolumntype{Y}{>{\raggedright\arraybackslash}X}

\begin{document}

\title{Q-BridgeNet: A Quantization Network for Cross-Lingual Sign Language Translation}

\titlerunning{Q-BridgeNet: A Quantization Network for Cross-Lingual SLT}

\author{Liqian Feng\inst{1}\and
Lintao Wang\inst{1} \and
Xiaochen Liu\inst{1} \and
Anusha Withana\inst{1} \and\\
Ken-Tye Yong\inst{1} \and
Dehui Kong\inst{2} \and
Zhiyong Wang\inst{1} \and
Kun Hu\inst{3}\textsuperscript{,\ding{41}}
}

\authorrunning{L. Feng et al.}

\institute{The University of Sydney, Darlington NSW, Australia\\
\email{\{lfen0902, lwan3720\}@uni.sydney.edu.au}\\
\email{\{xiaochen.liu, anusha.withana, ken.yong, zhiyong.wang\}@sydney.edu.au}\and
Beijing University of Technology, Beijing, China\\
\email{kdh@bjut.edu.cn}\and
Edith Cowan University, Joondalup WA, Australia\\
\email{k.hu@ecu.edu.au}
}

\begingroup
  \renewcommand\thefootnote{}
  \footnotetext{\textsuperscript{\ding{41}} Corresponding author.}
\endgroup

\maketitle

\begin{abstract}
  Most sign language translation (SLT) methods focus on isolated native sign–spoken pairs (e.g., American Sign Language–English). Extending language-specific SLT models to multilingual translation would improve accessibility by enabling communication across diverse sign and spoken language communities.
  However, existing multilingual SLT approaches still struggle to learn a unified model that minimizes cross-lingual conflicts while capturing shared cross-lingual semantics and preserving language-specific variations across different sign languages.
  Therefore, we propose \textbf{Q-BridgeNet}, a unified framework for multilingual SLT that jointly mitigates cross-lingual conflicts across both the sign language and spoken language sides. 
  On the sign language side, Q-BridgeNet learns discrete \emph{Q-units} via adaptive segmentation and residual vector quantization: a shared base codebook provides language-agnostic semantic primitives, while language-specific residual codebooks refine heterogeneous signing semantics.
  On the spoken language side, a multilingual LLM is fine-tuned to operate in the Q-unit space, leveraging cross-lingual priors to enable a unified SLT model.
  Experiments on PHOENIX14T, How2Sign, and CSL-Daily show that Q-BridgeNet effectively mitigates cross-lingual conflicts, achieving state-of-the-art performance on native sign–spoken pairs while also demonstrating strong generalization to non-native pairs. Our source code is publicly available at: \url{https://github.com/FengLiQ/Q-BridgeNet}
\end{abstract}

\section{Introduction}
\label{sec:intro}

\begin{figure*}[t]
  \centering
    \includegraphics[width=\linewidth]{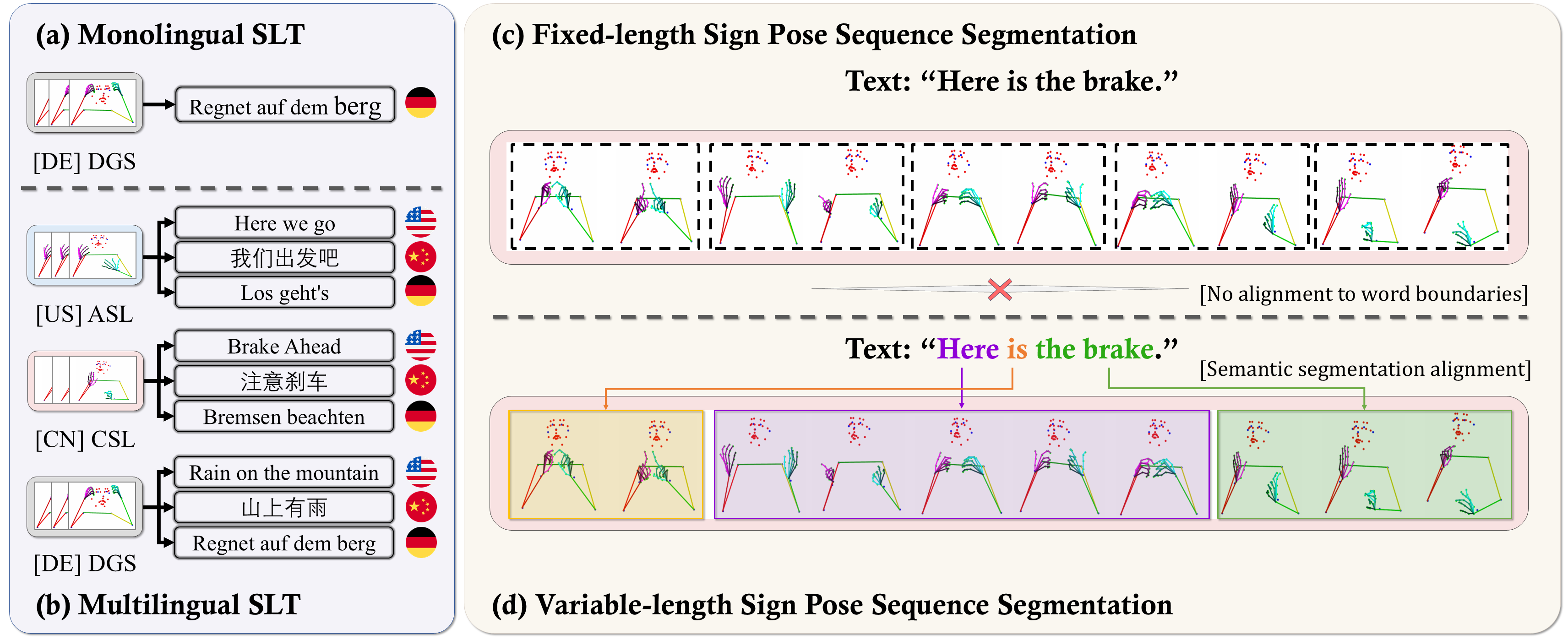}
    \caption{
    \textbf{Multilingual SLT and signing units.}
(a) Monolingual SLT translates a single sign--spoken language pair.
(b) Multilingual SLT extends to many-to-many translation across multiple sign and spoken languages.
(c) Fixed-length segmentation may cut across semantic/lexical boundaries, mixing heterogeneous signing patterns and yielding poor alignment to text.
(d) Variable-length semantic Q-Unit forms coherent signing units that better align with textual units and helps reduce cross-lingual interference.}
    \label{fig:fig_1}
\end{figure*}

Over 360 million people worldwide rely on sign languages for daily communication~\cite{nunez2023survey}. As visual languages with their own phonology, syntax, and discourse structure, sign languages differ fundamentally from spoken languages~\cite{Mann14012010}. Bridging the two is therefore a core challenge in multimodal language understanding, with direct implications for accessibility and inclusion.
Most prior work on \emph{Sign Language Translation} (SLT)~\cite{camgoz2018neural,camgoz2020sign,jiao2024visual,zhou2023gloss,lin-2023-gloss,gong2024llms} focuses on a \emph{single} sign-spoken language pair, learning to map continuous sign observations (e.g., video/pose sequences) into spoken-language text. 
These approaches have achieved substantial progress in monolingual SLT by learning mappings between visual sign representations and textual outputs.
While these methods have made strong progress in native (in-pair) settings, practical applications often require interaction across \emph{multiple} sign languages (\Cref{fig:fig_1} a,b) --e.g., American Sign Language (ASL), Chinese Sign Language (CSL), German Sign Language (DGS)--and \emph{multiple} spoken languages.

This motivates multilingual SLT, which can broaden accessibility by enabling communication across diverse sign and spoken language communities.
Recent efforts have begun to explore multilingual SLT~\cite{yin2022mlslt,UniSign,tan-etal-2025-multilingual}. 
However, these approaches still struggle to learn a \emph{unified} sign representation that aligns cross-lingual semantics while retaining language-specific realizations. 
When multiple sign languages are forced into a single shared visual embedding space, differences in visual realization can cause cross-lingual interference (negative transfer): language-specific patterns compete for representational capacity and corrupt the shared manifold, undermining robust multilingual modeling~\cite{shi2026culture}.

Therefore, we propose \textit{Q-BridgeNet}, a unified multilingual SLT framework that mitigates cross-lingual conflicts on \textit{both} the sign and spoken language sides via a shared–private residual quantization scheme. Our key idea is to decompose sign representations into  (i) \emph{language-agnostic semantic primitives} shared across sign languages and (ii) \emph{language-specific refinements} that account for linguistic variation.
Concretely, we first perform {adaptive temporal segmentation} (\Cref{fig:fig_1} c,d) to partition continuous sign motion into semantically coherent, variable-length units.
Each unit is then discretized with {residual vector quantization (RQ)}, where a {shared base codebook} captures cross-lingual primitives and {language-specific residual codebooks} model sign-language-specific variations. 
Finally, we fine-tune a {multilingual LLM} to operate directly on the resulting discrete \emph{Q-Unit} sequences, yielding a single many-to-many translation model across sign and spoken languages.

We evaluate our \textit{Q-BridgeNet} on three standard datasets--PHOENIX14T (DGS--German)~\cite{forster2014extensions}, How2Sign (ASL--English)~\cite{duarte2021how2sign}, and CSL-Daily (CSL--Chinese) \cite{huang2018video}--and construct many-to-many targets over German, English, and Chinese. \textit{Q-BridgeNet} achieves state-of-the-art performance on native pairs and demonstrates strong transfer to non-native pairs, approaching native-pair accuracy in several cross-lingual settings. 

Our main contributions are as follows:
\begin{itemize}
  \item \textbf{Q-BridgeNet.} We introduce a unified many-to-many multilingual SLT framework based on a shared--private residual quantization scheme.
  \item \textbf{Signing segmentation for Q-Unit discovery.} We propose a temporal segmentation mechanism that decomposes signing streams into semantically coherent variable-length units for discrete tokenization into \emph{Q-Units}.
  \item \textbf{LLM bridging in Q-unit space.} We fine-tune a multilingual LLM to operate in the discrete Q-Unit space for cross-lingual guidance.
\end{itemize}

\section{Related Work}
\label{sec:related_work}

\noindent\textbf{Sign Language Translation.} Early SLT methods typically adopt a two-stage pipeline, where sign videos were first transcribed into intermediate symbolic glosses labels before being translated into spoken text~\cite{camgoz2018neural, camgoz2020sign, kan2022sign, jiao2024visual}. 
However, gloss annotations can be costly and may act as an information bottleneck, which motivates recent gloss-free methods that directly translate sign into text. 
Gloss-free SLT approaches improve visual–linguistic alignment and representation learning through CTC-based modeling \cite{tan2025improvement}, dual-branch architectures \cite{chen2025c}, or multi-stream self-supervised pretraining \cite{gueuwou2025shubert}.
More recent methods increasingly leverage LLMs to strengthen cross-modal semantic grounding. 
MMSLT \cite{kim2025leveraging} integrates multimodal LLMs to enhance contextual understanding of sign motion and linguistic structure, while SpaMo \cite{hwang2025efficient} combines motion–spatial modeling with LLM-based semantic mapping for translation.
Although these advances improve SLT performance, they remain confined to single-language settings, limiting their ability to support communication across multiple sign and spoken language communities.

\noindent\textbf{Quantized Sign Representations.} 
Recent research \cite{jiang2023motiongpt,zhang2023generating, zhou2024avatargpt, yu2024signavatars,shen2024imagpose} has explored representing sign and motion sequences using vector-quantized models~\cite{wu2025dc} and motion tokenization approaches. And a data-driven representation\cite{walsh2024data} uses fixed-length VQ tokenization~\cite{shen2025imagdressing} for signs to map text to poses. \cite{gong2024llms} adopts LLMs for SLT by discretizing sign pose sequences via VQ-VAE. Yet, their tokenization uses fixed-length downsampling, limiting its ability to adapt to the variable temporal structure of sign language semantics. Moreover, the learned discrete tokens are usually constructed within a single-language setting and do not explicitly model cross-lingual semantic sharing across different sign languages. In contrast, our approach introduces hierarchical quantization with shared and language-aware residual codebooks built upon adaptive variable-length segmentation, enabling structured and transferable sign representations for multilingual SLT.

\noindent\textbf{Multilingual Sign Language Research.} 
Recent advances in sign languages have explored how models can generalize across multiple sign–spoken language pairs. 
MGSLT \cite{tan-etal-2025-multilingual} adapts LLMs for gloss-free multilingual SLT, translating sign features into several spoken languages through shared multimodal representations. 
IMSLT \cite{zhang-etal-2025-improving} improves cross-lingual transfer by clustering related sign languages and using these clusters as training priors. 
Signs as Tokens \cite{zuo2025soke} extends sign language production to a multilingual setting by treating signs as discrete motion tokens and augmenting generation with retrieved cross-lingual examples.
These works highlight an emerging shift toward unified multilingual sign language research that leverages linguistic similarity and shared visual patterns. 
Uni-Sign~\cite{UniSign} introduces a unified multilingual translation framework by leveraging shared visual encoders and multilingual language models to enable many-to-many sign–spoken translation. 
However, such approaches mainly rely on continuous visual representations and do not explicitly model structured discrete sign units that separate cross-lingual semantic primitives from language-specific variations.
Our approach Q-BridgeNet learns hierarchical discrete Q-units through shared–private quantization, enabling a structured sign representation that facilitates multilingual transfer.

\begin{figure*}[t]
  \centering
  \includegraphics[width=\linewidth]{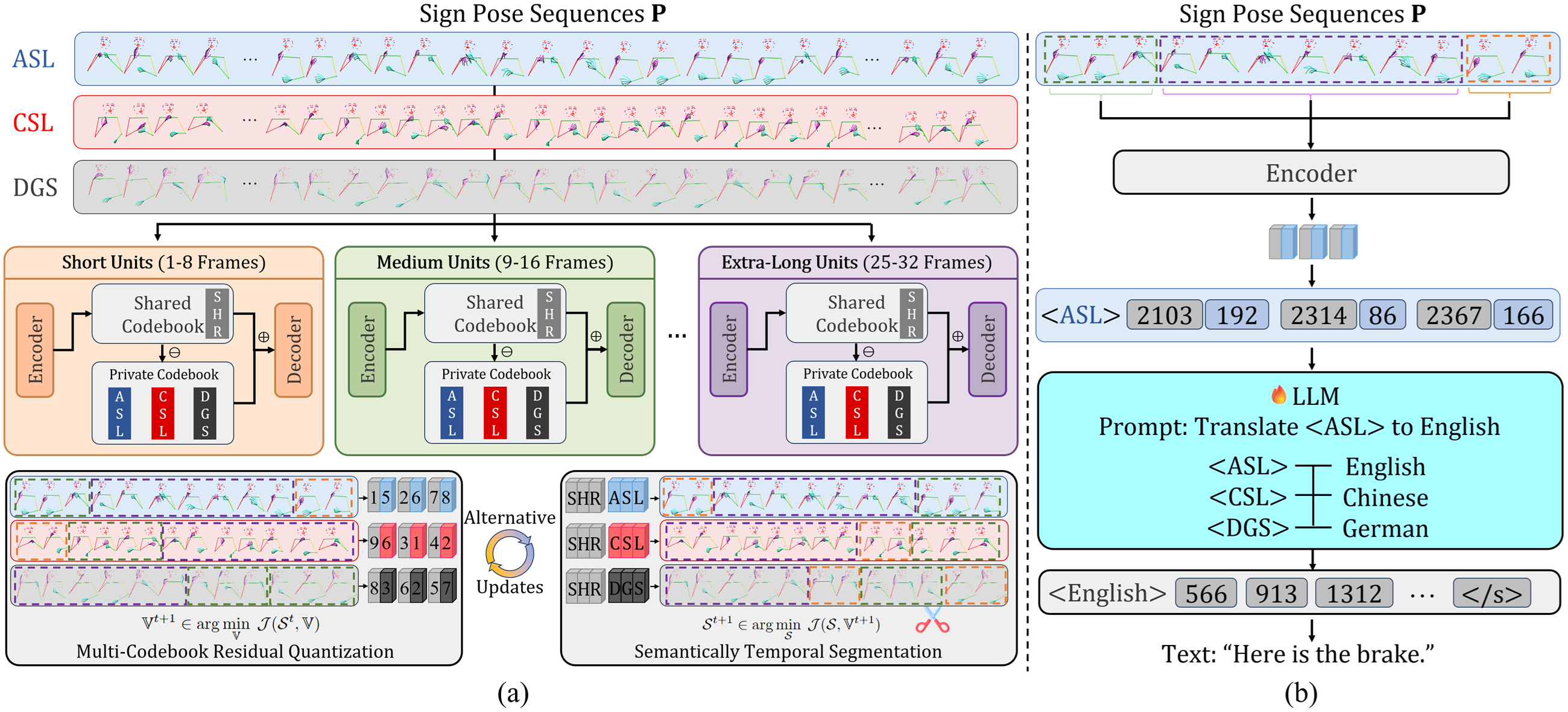}
  \captionof{figure}{{Q-BridgeNet Overview.}
(a) \emph{Iterative Q-Unit Discovery:} Segmentation alternates with residual VQ to produce variable-length discrete sign tokens from pose sequences. A \emph{shared base} codebook captures language-agnostic primitives, while \emph{language-specific} residual codebooks (ASL/CSL/DGS) add semantic detail.
(b) \emph{Multilingual LLM Re-Alignment:} A pretrained LLM is adapted to operate in the Q-unit space, enabling a unified \emph{many-to-many} SLT model across different sign and spoken languages.}
  \label{fig:methodology}
\end{figure*}

\section{Methodology}
\subsection{Overview}

\noindent\textbf{Architecture.} We introduce Q-BridgeNet (\Cref{fig:methodology}), a unified framework for many-to-many SLT that bridges visual signing and spoken language via a shared, discrete space. Any signing stream is represented as a sequence of quantized units (\textit{Q-units}), which serve as the atomic token for translation. Q-units are learned by alternating between (i) \textit{multi-codebook residual quantization} on variable‑length sign pose segments, featuring a shared base codebook for language-agnostic signing primitives and factor-binned residual codebooks for language-specific variation refinement and (ii) \textit{semantically temporal segmentation} of the pose stream based on previously trained residual codebooks. To connect Q-units to spoken languages, we fine-tune a pretrained multilingual LLM to operate in the Q-unit space, yielding a single model that supports both native (a sign language paired with its local spoken language) and cross-lingual (paired with a non-local spoken language) translation. 

\noindent\textbf{Notations.} In SLT, signing pose sequences and spoken language sentences are involved.  
We denote a signing pose sequence as  
$
\mathbf{P} = \{\mathbf{p}_1, \mathbf{p}_2, \dots, \mathbf{p}_T\},
$ with \( T \) frames, 
representing a signer’s motion trajectory. Each frame \( \mathbf{p}_t \) consists of \( K \) joints:  
$
\mathbf{p}_t = \{\mathbf{j}_{t,1}, \mathbf{j}_{t,2}, \dots, \mathbf{j}_{t,K}\},
$ 
where each joint \( \mathbf{j}_{t,k} \in \mathbb{R}^d \) stores its coordinates.
A signing clip is a contiguous subsequence of frames, denoted by the half-open interval
$
\mathbf{p}_{s:s'} = \{\mathbf{p}_s, \mathbf{p}_{s+1}, \dots, \mathbf{p}_{s'-1}\}
$ with $s<s'$. 
We denote the corresponding spoken-language sentence as
$
\mathbf{X} = (\mathbf{x}_1, \mathbf{x}_2, \dots, \mathbf{x}_M)
$
where $\mathbf{x}_m$ is the $m$-th word/token and $\mathbf{X}$ conveys the same meaning as $\mathbf{P}$.

\noindent\textbf{Signing sequence partitioning.}
To accommodate the irregular temporal dynamics of signing, we partition $\mathbf{P}$ into non-overlapping variable-length segments.
Formally, a segmentation is defined as
\[
\mathcal{S}(\mathbf{P}) = \{\mathbf{p}_{s_0:s_1}, \mathbf{p}_{s_1:s_2}, \dots, \mathbf{p}_{s_{N-1}:s_N}\},
\]  
where $1 = s_0 < s_1 < \dots < s_N = T+1$ are the segment boundaries.
This segmentation is later coupled with quantization to produce discrete Q-units, via a segment-to-codebook alignment mechanism detailed in subsequent sections.

\subsection{Residual Quantization for Multilingual Signing Unit}

Given a segmentation $\mathcal{S}(\mathbf{P})$, we discretize each sign  into a quantized unit (Q-units) using a shared--private residual VQ-VAE (RQ-VAE).

Our design combines (i) a \textit{shared base codebook} that captures language-agnostic signing primitives and 
(ii) \textit{language-specific private residual codebooks} that refine language-dependent variations (ASL/CSL/DGS).
This yields a compact discrete space that supports many-to-many learning across sign languages and spoken languages.

\noindent\textbf{RQ Encoding for Sign Segments. }
Formally, we define a set of RQ-VAE modules $\mathbb{V} = \{\mathcal{V}_1, \dots, \mathcal{V}_L\}$ for $L$ sign languages, where each $\mathcal{V}_i = \{\mathcal{E}, \mathcal{D}, \mathcal{Q}_i^{0:R}, \mathcal{Z}_i^{0:R}\}$ consists of an encoder, decoder, a stack of $R+1$ quantization layers with $R$ residuals, and their associated codebooks. 
Each module $\mathcal{V}_i$ handles segments with variable lengths for each sign language (ASL/CSL/DGS). 

Given a segment $\mathbf{p}_{s_n:s_{n+1}} \in \mathcal{S}(\mathbf{P})$, the encoder produces a latent representation $ \mathbf{z} = \mathcal{E}(\mathbf{p}_{s_n:s_{n+1}})$.
The quantization process further represents $\mathbf{z}$ with $R{+}1$ ordered quantization vectors:
\begin{equation}
\mathrm{RQ}(\mathbf{z}) = [\mathbf{z}^r]_{r=0}^{R},
\end{equation}
where $\mathbf{z}^r$ denotes the quantized code at the $r$-th residual layer. 
Starting from the initial residual $\mathbf{r}^0 = \mathbf{z}$, the quantization proceeds recursively as
\begin{equation}
\begin{aligned}
\mathbf{z}^r &= \mathcal{Q}_i^{r}(\mathbf{r}^r), \\
\mathbf{r}^{r+1} &= \mathbf{r}^r - \mathbf{z}^r,
\end{aligned}
\end{equation}
for $r = 0, \dots, R$.

\noindent\textbf{Q-Units. }
For cross-lingual representation, the first-layer quantizer $\mathcal{Q}^0_i$ and codebook $\mathcal{Z}^0_i$ are shared across all sign languages, reflecting the fact that the same meaning can span different lengths across sign languages. 
The residual quantizers $\{\mathcal{Q}^{r}_i\}_{r=1}^{R}$ employ \emph{factor-binned} codebooks $\{\mathcal{Z}_i^{r}\}$ that are language-specific; language variation is injected at this stage. 
This provides language-aware refinement of segment representations and supports many-to-many learning within a unified, adaptable latent space.
After the residual layer, the final quantized representation is the concatenation of all quantized residuals:
\begin{equation}
\hat{\mathbf{z}} = \mathbf{z}^0\oplus...\oplus\mathbf{z}^r\oplus...\oplus \mathbf{z}^R,
\end{equation}
which is then decoded as
\begin{equation}
\hat{\mathbf{p}}_{s_n:s_{n+1}} = \mathcal{D}_i(\hat{\mathbf{z}}).
\end{equation}

\noindent\textbf{Quantization Optimization. } The RQ-VAEs are optimized with a reconstruction and commitment loss:
\begin{equation}
\begin{aligned}
\mathcal{L}_{\text{VQ}} =\; & 
\| \mathbf{p}_{s_n:s_{n+1}} - \mathcal{D}_i(\hat{\mathbf{z}}) \|_2^2 \\
& + \beta \cdot \| \text{sg}[\mathbf{z}] - \hat{\mathbf{z}} \|_2^2
+ \gamma \cdot \| \mathbf{z} - \text{sg}[\hat{\mathbf{z}}] \|_2^2,
\end{aligned}
\end{equation}
where $\text{sg}[\cdot]$ denotes the stop-gradient operator. The first term ensures accurate reconstruction, while the latter enforces consistency between the encoding and selected code.

After residual quantization, the full signing sequence is represented as an ordered sequence of discrete latent tokens:
\begin{equation}
\hat{\mathbf{Z}} = (\hat{\mathbf{z}}_1, \hat{\mathbf{z}}_2, \dots, \hat{\mathbf{z}}_N).
\end{equation}
The resulting Q-units are compositional symbolic units across temporal scales, serving as atomic tokens for downstream translation and sign-text alignment.

\subsection{Joint Signing Segmentation-Quantization}

\noindent\textbf{Temporal Segmentation. }
Based on the learned RQ-VAE modules, we seek to construct the segmentation $\mathcal{S}(\mathbf{P})$. To this end, we define a global segmentation objective $\mathcal{L}_{\text{global}}$ that partitions the sequence into semantically coherent segments by minimizing the cumulative reconstruction error, where each segment is evaluated using an element-wise reconstruction loss $\mathcal{L}_{\text{elem}}$.
Formally, for a continuous signing sequence $\mathbf{p}_{s_0:s_N}$, the global segmentation loss is defined as:
\begin{equation}
\mathcal{L}_{\text{global}}(\mathbf{p}_{s_0:s_N} \mid \mathbb{V}) 
= \min_{\mathcal{S} = \{s_0, \dots, s_N\}} 
\sum_{n=0}^{N-1} \mathcal{L}_{\text{elem}}(\mathbf{p}_{s_n:s_{n+1}} \mid \mathbb{V}),
\end{equation}
where $\mathcal S$ denotes a valid segmentation of the input sequence. Each $\mathcal{L}_{\text{elem}}$ measures the reconstruction error of an individual segment reconstructed by the corresponding RQ-VAE module in $\mathbb{V}$:
\begin{equation}
\mathcal{L}_{\text{elem}}(\mathbf{p}_{s_n:s_{n+1}} \mid \mathbb{V}) = 
\| \mathbf{p}_{s_n:s_{n+1}} - \hat{\mathbf{p}}_{s_n:s_{n+1}} \|_2^2.
\end{equation}

Since the global segmentation involves a combinatorial search over all valid breakpoints, we adopt an approximation of the optimal segmentation. For a sequence ending at $s_N$, the optimal objective satisfies the following recurrence:
\begin{equation}
\label{equ:seg}
\begin{aligned}
\mathcal{L}_{\text{global}}(\mathbf{p}_{s_0:s_N} \mid \mathbb{V}) = \min_{s_{N-1}} \Big[
\mathcal{L}_{\text{global}}(\mathbf{p}_{s_0:s_{N-1}} \mid \mathbb{V}) \\
+ \mathcal{L}_{\text{elem}}(\mathbf{p}_{s_{N-1}:s_N} \mid \mathbb{V}) \Big].
\end{aligned}
\end{equation}
This recurrence decomposes the global optimization into a sequence of overlapping sub-problems. The base case is:
\begin{equation}
\mathcal{L}_{\text{global}}(\mathbf{p}_{0:1} \mid \mathbb{V}) = 
\mathcal{L}_{\text{elem}}(\mathbf{p}_{0:1} \mid \mathbb{V}).
\end{equation}
By solving all sub-problems sequentially, the optimal segmentation minimizing the cumulative reconstruction loss can be recovered through backtracking.

\noindent\textbf{Alternative Updates.} 
Segmentation and discrete tokenization are inherently coupled. Accurate temporal boundaries reduce intra-segment variation, allowing each segment to correspond to a more coherent semantic unit and thereby simplifying quantization, while a better learned discrete codebook provides more stable semantic prototypes, which in turn facilitates more reliable boundary estimation. Therefore, segmentation and quantization should be optimized jointly rather than independently. However, jointly optimizing the segmentation boundaries and discrete codebooks is difficult because both variables are interdependent.
We therefore formulate the following joint objective:
\begin{equation}
\begin{aligned}
\mathcal J(\mathcal S,\mathbb V)
& =\sum_{n=0}^{N-1}\!\Bigl[
\|\mathbf p_{s_n:s_{n+1}}-\mathcal D_i(\hat z_n)\|_2^2 \\
& +\beta\|\mathrm{sg}[z_n]-\hat z_n\|_2^2
+\gamma\|z_n-\mathrm{sg}[\hat z_n]\|_2^2
\Bigr]+\lambda N,
\end{aligned}
\end{equation}
where $\lambda N$ regularizes the number of segments.

Since directly optimizing $\mathcal J(\mathcal S,\mathbb V)$ is intractable, we adopt a block-coordinate descent strategy. The optimization is initialized from a randomly sampled valid segmentation, which serves as the starting point for the iterative refinement process. Each iteration alternates between updating the codebook and refining the segmentation, where the $t$-th update is as follows:

\noindent (A) {Codebook Update}: 
\begin{equation}
\mathbb V^{t+1} \in \arg\min_{\mathbb V}\ \mathcal J(\mathcal S^{t}, \mathbb V).
\end{equation}
(B) {Segmentation Update}: 
\begin{equation}
\mathcal S^{t+1} \in \arg\min_{\mathcal S}\ \mathcal J(\mathcal S, \mathbb V^{t+1}).
\end{equation}

Intuitively, each iteration first improves the representation quality under segmentation and then refines the segmentation according to the updated codebook, progressively aligning temporal boundaries with the learned semantic structure. This block-coordinate descent ensures monotonic decrease of $\mathcal J$, converging to a stable point and stopping when $(\mathcal J^{t}-\mathcal J^{t+1})/\mathcal J^{t}<\varepsilon$ or the change in breakpoints is below a threshold. The resulting segments are both temporally and semantically coherent, forming a discrete representation space well-suited for downstream SLT tasks.

\subsection{LLM Re-Alignment with Unified Latent Space}

To bridge signing and spoken language for downstream SLT, the trained RQ-VAE are frozen, an LLM is fine-tuned to align and adjust into the Q-unit space, leveraging its broad linguistic knowledge. 
For finetuning, we construct an extended multilingual, many-to-many sign-spoken dataset based on PHOENIX14T (DGS--German), How2Sign (ASL--English), and CSL-Daily (CSL--Chinese). 

For each signing sequence \( \mathbf{P} \), the original paired spoken sentence \( \mathbf{X}^{(\mathrm{la})} \) corresponds to its native spoken language \( \mathrm{la} \in \{\text{de}, \text{en}, \text{zh}\} \).  
To enable many-to-many supervision, we employ GPT-5 ~\cite{openai2025gpt5} to translate each native spoken caption \( \mathbf{X}^{(\mathrm{la})} \) to the remaining two languages and get a trilingual set  
\begin{equation}
\mathbb{X} = \{\mathbf{X}^{(\text{de})}, \mathbf{X}^{(\text{en})}, \mathbf{X}^{(\text{zh})}\}.
\end{equation}
Each signing sequence \( \mathbf{P} \) is paired with spoken-language sentences in all three languages, forming \((\mathbf{P}, \mathbb{X})\). 
This multilingual augmentation facilitates cross-lingual alignment and supports many-to-many learning across diverse sign and spoken modalities. 
Given a signing pose sequence, we segment and quantize it into a sequence of discrete tokens $\hat{\mathbf{Z}}$ via the frozen RQ-VAEs. The LLM is then finetuned to map between $\hat{\mathbf{Z}}$ and a corresponding spoken language sentence $\mathbf{X}\in\mathbb{X}$.

In SLT, the LLM is fine-tuned to generate $\mathbf{X}$ from $\hat{\mathbf{Z}}$ autoregressively:
\begin{equation}
\mathcal{L}_{\text{SLT}} = - \sum_{k=1}^{M} \log P(\mathbf{x}_k \mid \mathbf{x}_{<k}, \hat{\mathbf{Z}}).
\end{equation}
This unified next-token prediction design enables translation between multiple signing and natural language, grounded in a shared latent space and facilitated by the LLM’s semantic prior.

\section{Experiments}

\subsection{Experimental Settings}

\noindent\textbf{Dataset.} We evaluate our method on a multilingual corpus aggregated from three widely used sign language datasets: PHOENIX14T \cite{forster2014extensions}, How2Sign \cite{duarte2021how2sign}, and CSL-Daily \cite{huang2018video}. 
PHOENIX14T contains approximately 8K German Sign Language (DGS) clips from weather forecast broadcasts. How2Sign consists of around 35K American Sign Language (ASL) samples extracted from instructional videos featuring diverse signers and content. CSL-Daily offers roughly 20K Chinese Sign Language (CSL) samples of daily conversations, supporting evaluation under a low-resource, non-western setting. Following prior work, we adopt the standard evaluation protocol defined for each dataset to ensure fair comparison with existing methods.

\definecolor{ColorA}{HTML}{F5276C} 
\definecolor{ColorB}{HTML}{188DB4}

\definecolor{ColorA}{HTML}{F5276C} 
\definecolor{ColorB}{HTML}{188DB4}
\begin{table}[t]
  \setlength{\tabcolsep}{1pt}
  \renewcommand{\arraystretch}{0.9}
  \centering
  \footnotesize
  \begin{tabularx}{\linewidth}{@{}l >{\centering\arraybackslash}X >{\centering\arraybackslash}X >{\centering\arraybackslash}X |l >{\centering\arraybackslash}X >{\centering\arraybackslash}X >{\centering\arraybackslash}X |l >{\centering\arraybackslash}X >{\centering\arraybackslash}X >{\centering\arraybackslash}X @{}}
    \toprule
    \multirow{2}{*}{\textbf{Method}}  & \multicolumn{3}{c|}{\cellcolor{gray!20}\textbf{How2Sign}} & \multirow{2}{*}{\textbf{Method}} &  \multicolumn{3}{c|}{\cellcolor{gray!20}\textbf{CSL-Daily}} & \multirow{2}{*}{\textbf{Method}} & \multicolumn{3}{c}{\cellcolor{gray!20}\textbf{PHOENIX14T}} \\
    & \textbf{B-4} & \textbf{B-1} & \multicolumn{1}{c|}{\textbf{RG}} & & \textbf{B-4} & \textbf{B-1} &  \multicolumn{1}{c|}{\textbf{RG}} & & \textbf{B-4} & \textbf{B-1} & \textbf{RG} \\
    \midrule
    SL-Trans.    & 9.93 & 28.37 & 28.94 &SL-Trans. & 14.15 & 39.72 & 38.44&SL-Trans. & 19.45 & 43.72 & 45.44\\
    Uni-Sign& 14.50 & \secondbest{40.40} & \secondbest{34.30}&Uni-Sign& \secondbest{25.61} & 53.86 & \secondbest{54.92}&TwoStr.& 28.42 & 54.22 & 53.19\\
    GloFE    & 2.24 & 14.94 & 12.61 & MSLU & 11.42 & 33.97 & 33.80&CV-SLT& 29.27 & 54.88 & 54.33\\ 
    YT-ASL   & 12.39 & 37.82 &--&MSKA& 25.52 & \secondbest{56.37} & 54.04 &MSKA& 29.03 & 54.79 & 53.54\\
    ShuBert& \secondbest{16.20}&--&--&MMSLT& 21.11 & 49.87 & 48.92 &MMSLT & 25.73 & 48.92 & 47.97\\
    &&&&ICTCA & 22.47 & 52.37 & 51.87 & ICTCA & 28.42 & 53.95 & 53.34\\
    &&&&&&&&SCOPE& \secondbest{32.84} & \secondbest{61.74} & \secondbest{60.06} \\
    \midrule
    \multicolumn{4}{l}{\textbf{Ours} \textcolor{ColorB}{(Multilingual)}}\\
    Avg.& 16.44 & 41.73 & \secondbest{35.20}&Avg. & 26.33 & \secondbest{57.09} & \secondbest{55.04}&Avg. & \secondbest{33.40} & 62.56 & 61.39\\
    ASL-EN$\clubsuit$ & \secondbest{16.55} & \secondbest{41.81} & 34.97&CSL-EN&\secondbest{26.40} & 56.98 & 54.30&DGS-EN&33.20 & 62.21 & 60.73 \\
    ASL-ZH & \best{16.83} & \best{42.04} & \best{35.52}&CSL-ZH$\clubsuit$ &\best{26.91} & \best{57.39} & 54.85&DGS-ZH& 33.36 & \secondbest{62.67} & \best{61.67} \\
    ASL-DE & 15.96 & 41.34 & 35.11&CSL-DE&25.68 & 56.89 & \best{55.97}&DGS-DE$\clubsuit$& \best{33.62} & \best{62.80} & \secondbest{61.76}\\
    \midrule
    \multicolumn{4}{l}{\textbf{Ours} \textcolor{ColorA}{(Monolingual)}$\clubsuit$}\\
    ASL-EN$\clubsuit$ & 14.35 & 37.81 & 32.77 &CSL-ZH$\clubsuit$& 24.08 & 52.37 & 50.27 &DGS-DE$\clubsuit$& 30.16 & 56.02 & 53.75\\  
    \bottomrule
  \end{tabularx}
  \caption{Performance comparison on the SLT task. $\clubsuit$ indicates the native sign-spoken language pair in the dataset.}
  \label{tab:performance__slt}
\end{table}

\noindent\textbf{Implementation Details.} 
Following existing studies, {pose/skeleton} pipelines (landmarks of hands–body–face) are a prevalent choice, as they preserve articulatory structure while reducing appearance bias and privacy risk. We represent each sign pose frame $\mathbf{p}_i$ using 79 keypoints (21 per hand, 11 for body, 26 for face), with a maximum sequence length $T=256$ for efficiency.
For preliminary segmentation, we limit each segment to maximum 32 frames.
For multi-codebook design, we employ three language-specific RQ-VAEs, with a 96-dimensional latent space, corresponding to the ASL, CSL, DGS datasets. The shared codebook across these three RQ-VAEs consists of 1,024 embeddings, each RQ-VAE maintains a private codebook of 512 embeddings. 
We train the RQ-VAEs for 1,000 epochs with hyperparameters $\beta\!=\!1$, $\gamma\!=\!0.25$ empirically.
Segmentation and quantization are co-optimized in an alternating manner, which is repeated for 5 times and the initial segmentation is created randomly.
For LLM realignment, we fine-tuned Qwen3-1.7B~\cite{Qwen-IG} for both SLT using a shared prompting schema with LoRA~\cite{hu2022lora}. Sign tokens are tagged with language labels and paired with their corresponding spoken texts. Prompts like \underline{Translate $<$DGS$>$ to German} guide the sign language translation task. 
We finetune for 100 epochs across the extended multilingual dataset on four NVIDIA RTX 4090 GPUs.
To enable many-to-many supervision, we translate each native spoken caption into the remaining two languages using GPT-5\cite{openai2025gpt5}, forming a trilingual caption set for every sign instance. Furthermore, As the keypoints in datasets are extracted from raw video frames using human pose estimation methods (e.g., OpenPose~\cite{OpenPose}) and these pose estimation methods are inherently imperfect, both training and test data contain noisy keypoint measurements. This aligns with practical scenarios in real-world applications, where visual modality data inevitably carries noise due to occlusions, varying lighting conditions, and estimation errors. To improve robustness and simulate realistic variations, we apply standard data augmentation techniques and additionally inject i.i.d. Gaussian noise with standard deviation 0.002 to the normalized keypoint coordinates during training. Supplementary Material provides more details. 

\noindent\textbf{Evaluation Metrics.}
For evaluation, we follow standard protocols established in prior SLT work. 
We assess the quality of generated spoken language using BLEU~\cite{papineni2002bleu,post-2018-call} and ROUGE~\cite{lin2004rouge}, two widely adopted metrics for machine translation~\cite{tan-etal-2025-multilingual}. For native SLT evaluation, we report results against the original ground-truth of each benchmark. For cross-lingual sign–spoken pairs where no original ground-truth references exist, we evaluate against the corresponding GPT-5\cite{openai2025gpt5} translated captions.

\noindent\textbf{Baselines.}
For baselines, we compare our method against existing state-of-the-art SLT models, including SL-Transformers \cite{camgoz2020sign}\src{CVPR'20}, TwoStreamSLT \cite{chen2022two}\src{NIPS'22}, GloFE\cite{lin-2023-gloss}\src{ACL'23}, YT-ASL \cite{uthus2023youtube}\src{NIPS'23}, CV-SLT \cite{zhao2024conditional}\src{AAAI'24},MSLU \cite{zhou2025scaling}\src{TPAMI'24}, MSKA \cite{guan2025mska}\src{PR'25}, SCOPE \cite{liu2025scope}\src{AAAI'25}, MMSLT\cite{kim2025leveraging}\src{ICCV'25}, ICTCA\cite{tan2025improvement}\src{COLING'25}, ShuBert \cite{gueuwou2025shubert}\src{ACL'25}, and Uni-Sign\cite{UniSign}\src{ICLR'25}.

\begin{table}[t]
  \setlength{\tabcolsep}{1pt}
  \renewcommand{\arraystretch}{0.9}
  \centering
  \footnotesize
  \begin{tabularx}{\linewidth}{@{}l >{\centering\arraybackslash}X >{\centering\arraybackslash}X >{\centering\arraybackslash}X >{\centering\arraybackslash}X >{\centering\arraybackslash}X >{\centering\arraybackslash}X >{\centering\arraybackslash}X >{\centering\arraybackslash}X >{\centering\arraybackslash}X @{}}
    \toprule
    \multirow{2}{*}{\textbf{Method}}  & \multicolumn{3}{c|}{\cellcolor{gray!20}\textbf{How2Sign}} & \multicolumn{3}{c|}{\cellcolor{gray!20}\textbf{CSL-Daily}} & \multicolumn{3}{c}{\cellcolor{gray!20}\textbf{PHOENIX14T}} \\
    & \textbf{B-4} & \textbf{B-1} & \multicolumn{1}{c|}{\textbf{RG}} & \textbf{B-4} & \textbf{B-1} &  \multicolumn{1}{c|}{\textbf{RG}} & \textbf{B-4} & \textbf{B-1} & \textbf{RG} \\
    
    \midrule
    
    \textbf{Ours}                  & \best{16.44} & \best{41.73} & \best{35.20}& \best{26.33} & \best{57.09} & \best{55.04}& \best{33.40} & \best{62.56} & \best{61.39} \\
    Uni-codebook VQ    & 7.23 & 21.61 & 18.09 & 16.18 & 33.97 & 35.12& 21.38 & 42.34 & 39.22 \\
    Non-shared VQ              & 9.29 & 25.60 & 22.52& 18.90 & 37.82 & 39.01 & 22.62 & 42.03 & 43.14  \\
    Non-residual VQ            & 12.04 & 31.48 & 27.75& 21.49 & 41.44 & 44.32  & 27.32 & 50.98 & 52.45 \\
    Single-pass Seg. \& RQ & 13.39 & 33.24 & 31.56 & 24.11 & 46.47 & 47.36 & 28.61 & 52.03 & 49.46\\
    GPT-5 Translation &  14.20 & 36.61 & 31.82 & 23.18 & 51.47 & 50.02 & 28.36 & 54.13 & 52.26\\
    GPT-4.1 Created Caption & 15.58 &  \secondbest{41.36} & 34.07 & \secondbest{26.30} & \secondbest{55.92} & \secondbest{54.48} & \secondbest{33.07} & \secondbest{62.34} & \secondbest{60.79}  \\
    Qwen3-0.6B Fine-Tuning & \secondbest{15.64} & 41.01 & \secondbest{34.20} & 25.84 & 55.41 & 53.54 & 32.25 &60.76 & 59.01 \\

    \bottomrule
  \end{tabularx}
  \caption{Ablation study results for SLT.}
  \label{tab:ablation1}
\end{table}

\begin{figure*}[t]
  \centering
    \includegraphics[width=\linewidth]{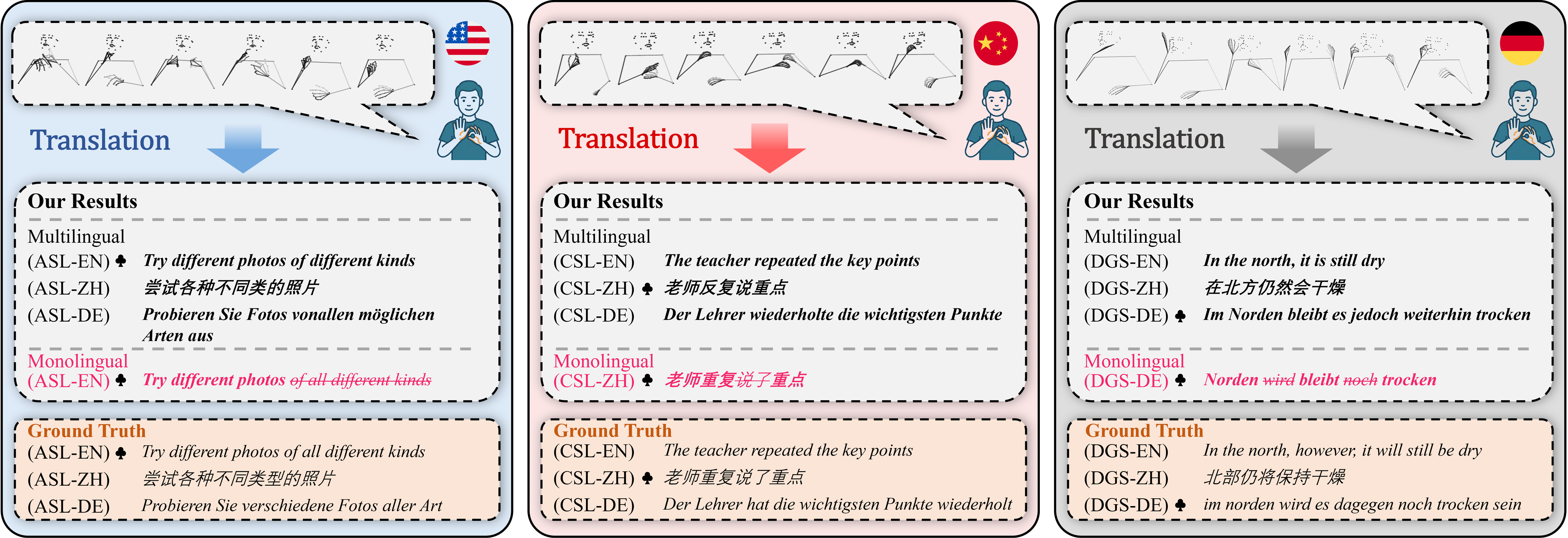}
    \\[1mm]
    \includegraphics[width=\linewidth]{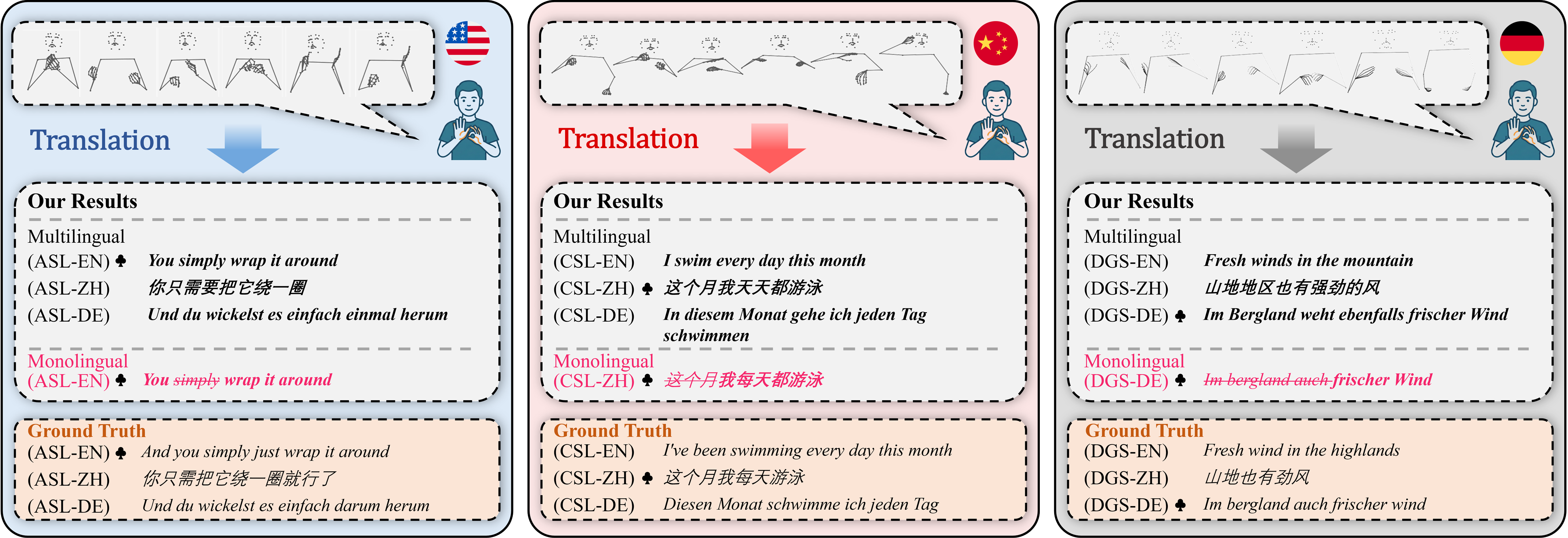}
    \caption{Multilingual qualitative examples for SLT. \textbf{Bold}: semantically aligned with GT; and \st{Strike}:  missing semantics.}
    \label{fig:qualitative1}
\end{figure*}

\subsection{Comparison with State-of-the-Art}

Our method achieves state-of-the-art results across all evaluation metrics for SLT on \textit{native} PHOENIX14T, How2Sign, and CSL-Daily, demonstrating the robustness and versatility of our unified framework.

Q-BridgeNet establishes a consistent lead on \emph{BLEU-4} across all three benchmarks, with clear margins over the strongest prior on each dataset. Crucially, these gains do not come at the expense of surface precision: wherever baselines report \emph{BLEU-1} and \emph{ROUGE}, Q-BridgeNet is also best, indicating that the improvements extend beyond long $n$-grams to token-level accuracy and summary overlap. The advantage holds against diverse families (seq2seq, multi-stream, alignment-oriented, and recent large-model pipelines), suggesting robustness to architectural choices.
Q-units, formed by adaptive segmentation and residual VQ, provide a more stable discrete interface for the LLM to compose multi-token semantics, especially when signing tempo and segment length distributions differ across datasets. 

\subsection{Cross-Lingual Non-Native Performance}
Most SLT benchmarks evaluate only native sign–spoken pairs, leaving cross-lingual generalization largely unexplored. Q-BridgeNet uniquely enables and excels in such non-native settings, demonstrating true many-to-many communication capability.

Across all three corpora, the multilingual model outperforms its monolingual counterpart trained only on the native pair, indicating that cross-lingual supervision regularizes the discrete signing space and improves potential lexical/phrase coverage.
Strikingly, the native target is \emph{not} always optimal: on DGS and ASL, non-native targets (e.g., ASL$\rightarrow$zh) edge out the native ones, suggesting that a shared Q-unit inventory with multilingual decoding can recover semantics underrepresented in native captions—or easier to learn in certain target languages. 
On CSL, the native target remains the best single direction, yet the multilingual average still exceeds monolingual training, pointing to a net benefit from many-to-many constraints even when the native data are strong.
Overall, these trends support our design goal: language-agnostic Q-units enable transfer across spoken targets without re-learning sign units per pair.

\subsection{Qualitative Evaluation}
We present representative examples in \Cref{fig:qualitative1} for SLT, highlighting both native and non-native translation scenarios. Q-BridgeNet consistently preserves the core semantics of the input sign sequences across English, Chinese, and German, demonstrating the effectiveness of our shared–private Q-unit representation in capturing cross-lingual semantic primitives while maintaining language-specific nuances. 

In native settings, the model produces fluent and accurate translations that align closely with the ground-truth text, effectively leveraging both the shared base codebook and language-specific residual codebooks to handle linguistic variations. In non-native scenarios—where sign languages are paired with a spoken language not seen during training—Q-BridgeNet still maintains semantic coherence, often producing translations that retain the meaning of the source signs, whereas a monolingual ablation baseline frequently misses key information or produces incomplete translations. 

These observations suggest that our variable-length semantic segmentation, combined with hierarchical quantization, not only generates semantically meaningful discrete units but also provides a structured representation that can be effectively interpreted by the multilingual LLM. Overall, this evaluation illustrates the benefits of unified many-to-many modeling and highlights the potential of Q-unit representations to facilitate robust cross-lingual transfer in SLT.

\subsection{Ablation Study}
We analyze variants to demonstrate the effectiveness of the proposed mechanisms. Results are reported in \Cref{tab:ablation1}.

\noindent\textbf{Uni-Codebook VQ (Vanilla VQ-VAE).}
One global codebook is used for all segments regardless of the language. 
This variant is consistently the weakest across datasets and tasks. BLEU-4 collapses (e.g., PHOENIX14T from {33.40} to 21.38).  
A single codebook is too coarse to capture multi-lingual complexity; language-specific factors should be modeled explicitly in a many-to-many setting.

\noindent\textbf{Non-Shared VQ (No Shared Base).}
Remove the shared base codebook; each language bin learns its own first-layer codebook.  
Performance improves over single–codebook on SLT, while remains behind the full model. The latent space becomes fragmented across bins, limiting cross-bin/unit multilingual reuse. 
Without a language-agnostic pivot, units learned in different bins fail to align semantically, hindering transfer and compositionality.

\noindent\textbf{Non–Residual VQ (Flat Multi-Codebook, No Residual Stack).}
Replace residual stacking with parallel (concatenated) codebooks per language bin, akin to multi-codebook VQ-VAE. This is a mid-point: it recovers a large portion of SLT text performance, but still lags the full design on all three datasets. The gap is stable across metrics.  
Language-specific factorization helps, but without residual “error peeling,” the representation lacks the fine-grained refinement provided by stacked residuals. 

\noindent\textbf{Single-pass Segmentation \& Quantization.} Using the full residual quantizer but reduce to a single segmentation–quantization pass lags behind the full model, as it lacks the boundary–code feedback: iterative refinement reconciles mismatches between initial segments and token assignments. In short, single-pass+RQ is a strong first cut; alternating refinement closes the remaining gap. 

\noindent\textbf{GPT-5 Translation.} To verify whether the multilingual gains come from cross-lingual sign modeling rather than GPT-5~\cite{openai2025gpt5} translation, we construct a control baseline that first performs monolingual SLT and then translates the predicted output into non-native target languages using GPT. Although this baseline is evaluated against the same GPT-generated references, it does not perform cross-lingual sign modeling. As shown in \Cref{tab:ablation1}, its consistently lower performance suggests that the multilingual gains stem from learning cross-lingual sign representations rather than GPT-based post-hoc translation.

\noindent\textbf{GPT-4.1 Created Caption.} To assess the effect of translation quality in multilingual supervision, we replace GPT-5~\cite{openai2025gpt5} with GPT-4.1~\cite{openai2025gpt4} while keeping all other settings unchanged. GPT-4.1 yields slightly lower SLT performance than GPT-5, indicating that higher-quality translations provide more reliable multilingual supervision. Nevertheless, the results remain comparable, suggesting that the proposed framework is not overly sensitive to moderate variations in translation quality. 

\noindent\textbf{Qwen3-0.6B Fine-Tuning.} To evaluate the effect of LLM scale, we replace Qwen3-1.7B~\cite{Qwen-IG} with the lightweight Qwen3-0.6B while keeping the remaining configurations fixed. As shown in \Cref{tab:ablation1}, the smaller backbone leads to only a modest degradation across all evaluation metrics. The consistently competitive performance indicates that the proposed Q-unit representation generalizes well across different LLM scales, making the framework suitable for resource-constrained deployment.

\begin{figure*}[t]
  \centering
    \includegraphics[width=\linewidth]{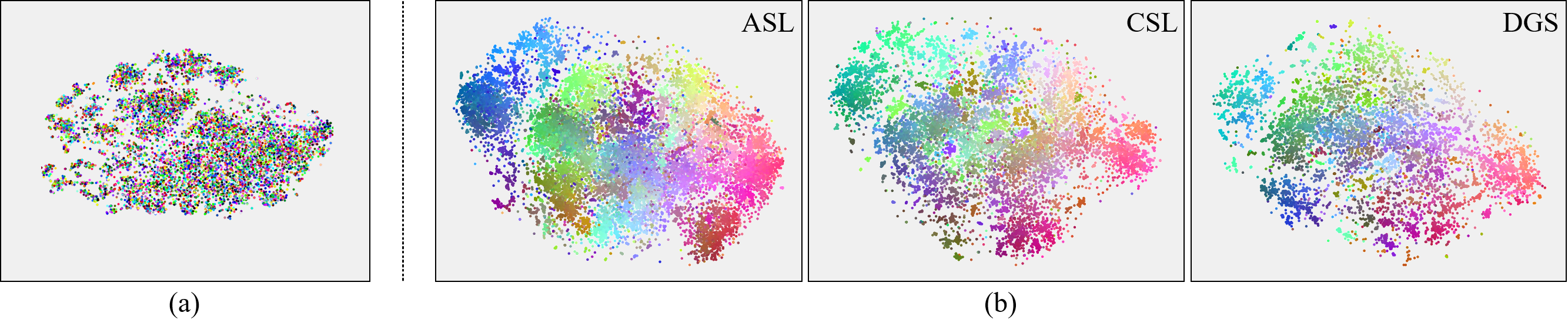}
    \caption{t-SNE visualization of discrete sign tokens. (a) Fixed-length segmentation with a single VQ-VAE results in scattered and semantically entangled clusters. (b) Our variable-length segmentation with three language-specific codebooks ($\mathcal{V}_{ASL}$, $\mathcal{V}_{CSL}$, and $\mathcal{V}_{DGS}$) yields well-separated clusters, where similar motion units are grouped and color-coded by semantic similarity.}
    \label{fig:fig2}
\end{figure*}

\begin{figure*}[t]
  \centering
    \includegraphics[width=\linewidth]{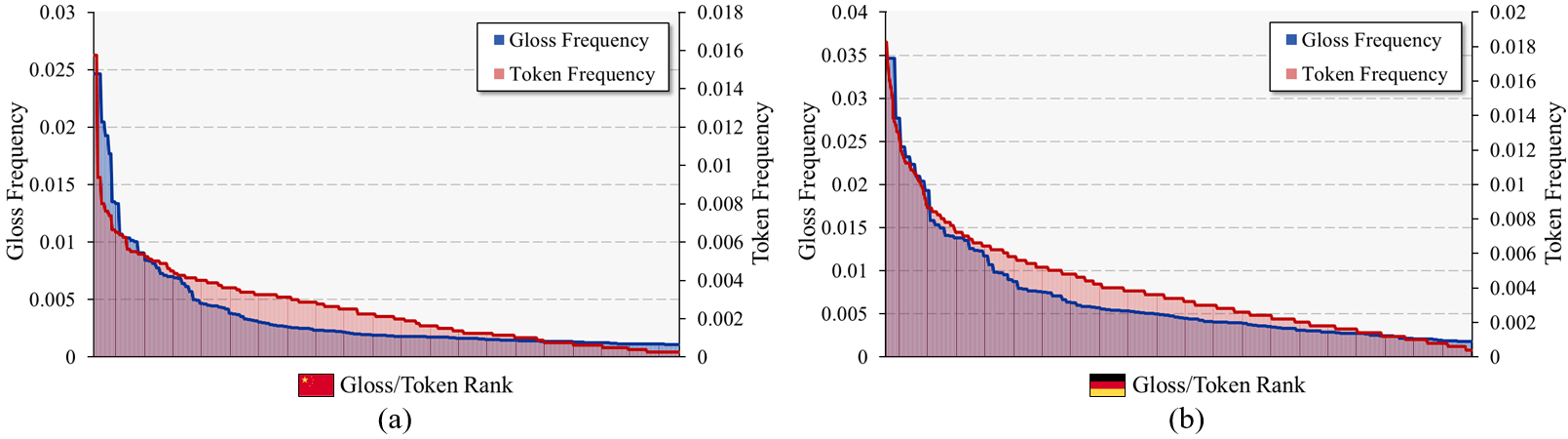}
    \caption{Comparison between gloss frequency distributions (blue) and private Q-unit token distributions (red) on CSL and DGS. The similar long-tailed trends indicate structural resemblance between learned Q-units and linguistic-level annotation units.
}
    \label{fig:frequency}
\end{figure*}

\subsection{Analysis of Q-unit Space Distribution}

To further examine the structural properties of the learned Q-unit space, we visualize segment-level latent representations using t-SNE \cite{maaten2008visualizing}. 
Each motion segment is projected into a 2D space and color-coded based on its assigned discrete token.

We compare our multi-codebook design—based on variable-length segmentation and language-specific codebooks ($\mathcal{V}_{ASL}$, $\mathcal{V}_{CSL}$, and $\mathcal{V}_{DGS}$)—with a baseline that employs fixed-length segmentation and a single codebook, consistent with our ablation setup. 
As shown in \Cref{fig:fig2}, the baseline exhibits a relatively scattered distribution with limited cluster structure, suggesting weaker organization in the learned latent space. 

In contrast, our approach produces more compact and visually separable clusters, where segments assigned to the same discrete token tend to group together. 
This improved spatial coherence indicates that the hierarchical segmentation and shared–private quantization strategy encourages structurally organized and semantically consistent representations.

To investigate whether the learned Q-units exhibit meaningful structural properties, we analyze their frequency distributions and compare them with gloss annotations available in the CSL and DGS datasets. The ASL dataset does not provide gloss annotations and is therefore excluded from this comparison.

Gloss annotations are commonly regarded as linguistic-level symbolic units that reflect semantic decomposition in sign language corpora. As shown in \Cref{fig:frequency}, we compare the frequency distributions of gloss tokens with those of private Q-unit tokens learned by our model. Specifically, the x-axis represents gloss words and language-specific Q-unit tokens sorted by descending frequency, while the left y-axis shows the occurrence frequency of gloss words and the right y-axis shows the occurrence frequency of Q-unit tokens. Both glosses and Q-unit tokens exhibit clear long-tailed distributions, and their decay trends are structurally similar across both CSL and DGS datasets.

Although Q-units are learned without gloss supervision, their usage patterns resemble those of manually annotated gloss units in terms of distributional structure. This suggests that the hierarchical quantization mechanism captures reusable structural patterns rather than arbitrary discretization artifacts. Importantly, this similarity emerges purely from data-driven segmentation and residual quantization, indicating that the learned token space exhibits semantic compactness and organized structure.

These observations support our design choice of modeling sign language through hierarchical discrete units and validate the effectiveness of the shared-private decomposition in learning transferable yet expressive representations.

\section{Conclusion} 
We presented \textbf{Q-BridgeNet}, a unified multilingual SLT framework that mitigates cross-lingual conflicts across sign and spoken languages by learning discrete \emph{Q-units} via adaptive variable-length segmentation and hierarchical shared–private residual vector quantization, capturing both cross-lingual semantic primitives and language-specific variations. A multilingual LLM is fine-tuned to operate in the Q-unit space, enabling a single many-to-many SLT  model. Experiments on PHOENIX14T, How2Sign, and CSL-Daily demonstrate state-of-the-art performance on native pairs and strong generalization to non-native pairs.

While this work focuses on multilingual many-to-many SLT, a promising direction is to extend the framework to cross-lingual sign language production (SLP), where spoken language inputs are mapped back to sign sequences. Since Q-units provide a structured and transferable discrete sign space, a unified Q-unit representation could serve as a bidirectional bridge between multilingual spoken and sign languages, enabling scalable and symmetric sign–text modeling.

\section*{Acknowledgements}
This research was supported by the Australian Research Council (ARC) under Project LP230100294 and the Edith Cowan University (ECU) Early-Mid Career Researcher (EMCR) Grant Scheme. Dr. Anusha Withana is the recipient of a Future Fellowship (FT250100813) funded by ARC.

\bibliographystyle{splncs04}
\bibliography{main}
\end{document}